\documentclass[10pt,twocolumn,letterpaper]{article}

\usepackage{cvpr}
\usepackage{times}
\usepackage{epsfig}
\usepackage{graphicx}
\usepackage{amsmath}
\usepackage{amssymb}
\usepackage{epstopdf}
\usepackage{multirow}
\usepackage{dsfont}
\usepackage{amsfonts}
\usepackage{booktabs}
\usepackage{threeparttable}
\usepackage{color}
\usepackage{stfloats}
\usepackage{bm}

\usepackage[pagebackref=true,breaklinks=true,letterpaper=true,colorlinks,bookmarks=false]{hyperref}

\cvprfinalcopy 

\def\cvprPaperID{****} 

\ifcvprfinal\fi
\begin{document}

\title{Iterative Context-Aware Graph Inference for Visual Dialog}

\author{Dan Guo$^{1,2}$\quad Hui Wang$^{1,2*}$\quad Hanwang Zhang$^3$\quad Zheng-Jun Zha$^4$\quad Meng Wang$^{1,2}$\thanks{Corresponding authors.}\\
$^1$Key Laboratory of Knowledge Engineering with Big Data, Hefei University of Technology\\
$^2$School of Computer Science and Information Engineering, Hefei University of Technology\\
$^3$Nanyang Technological University\\
$^4$University of Science and Technology of China\\
{\tt\small guodan@hfut.edu.cn, \{wanghui.hfut, eric.mengwang\}@gmail.com,}\\
{\tt\small hanwangzhang@ntu.edu.sg, zhazj@ustc.edu.cn}
}

\maketitle
\thispagestyle{empty}

\begin{abstract}
    Visual dialog is a challenging task that requires the comprehension of the semantic dependencies among implicit visual and textual contexts. This task can refer to the relation inference in a graphical model with sparse contexts and unknown graph structure (relation descriptor), and how to model the underlying context-aware relation inference is critical. To this end, we propose a novel Context-Aware Graph (CAG) neural network. Each node in the graph corresponds to a joint semantic feature, including both object-based (visual) and history-related (textual) context representations. The graph structure (relations in dialog) is iteratively updated using an adaptive top-$K$ message passing mechanism. Specifically, in every message passing step, each node selects the most $K$ relevant nodes, and only receives messages from them. Then, after the update, we impose graph attention on all the nodes to get the final graph embedding and infer the answer. In CAG, each node has dynamic relations in the graph (different related $K$ neighbor nodes), and only the most relevant nodes are attributive to the context-aware relational graph inference. Experimental results on VisDial v0.9 and v1.0 datasets show that CAG outperforms comparative methods. Visualization results further validate the interpretability of our method.
\end{abstract}
\section{Introduction}
\begin{figure}[t]
\centering
  \includegraphics[width=8cm]{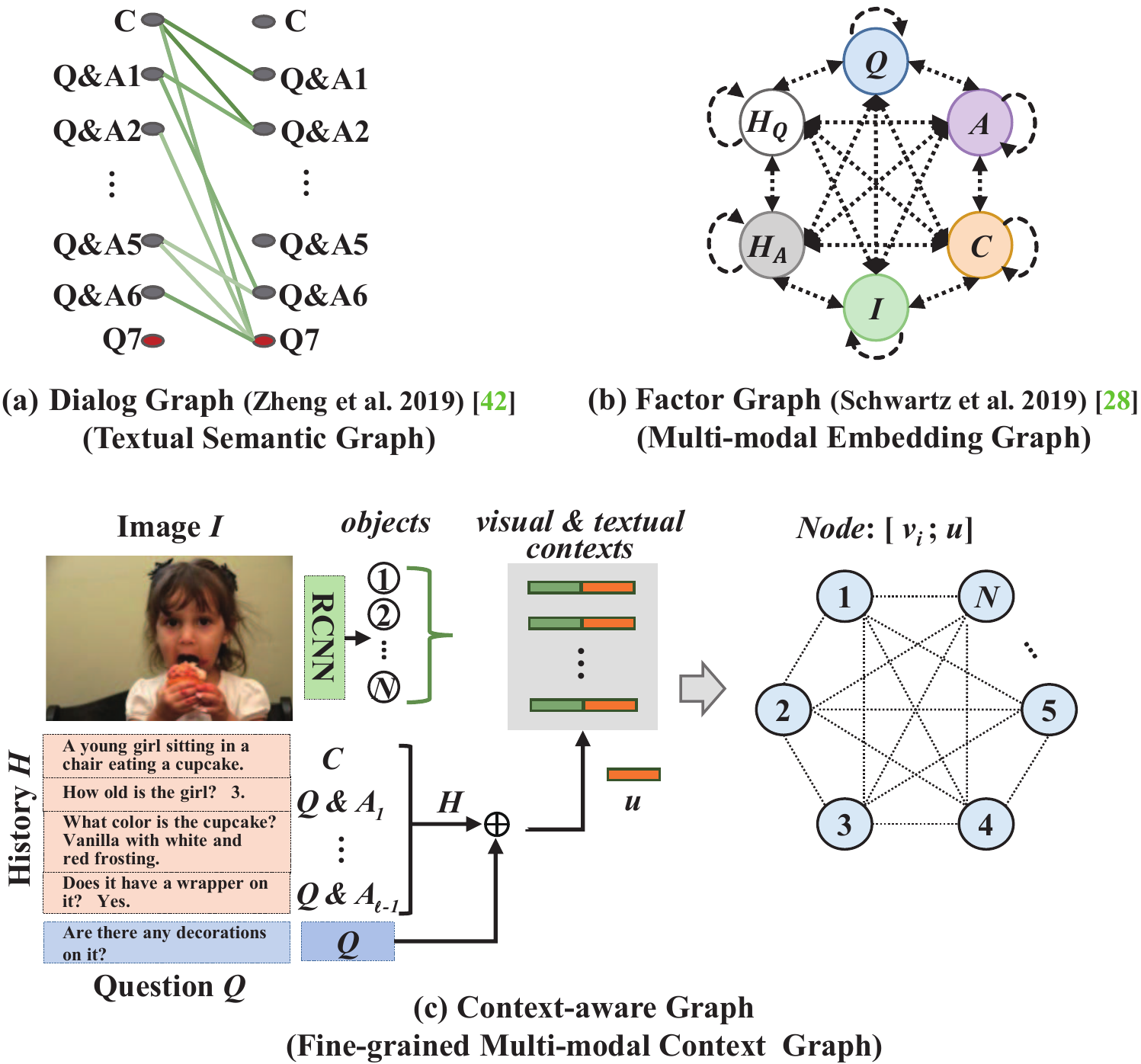}\\
  \caption{Different graph structures for visual dialog. In our solution (c), we focus on a question-conditioned context-aware graph, including both fine-grained visual-objects and textual-history semantics.
  }\label{figure1}
\end{figure}
Recently, cross-modal semantic understanding between vision and language has attracted more and more interests, such as image captioning~\cite{DBLP:conf/icml/XuBKCCSZB15,DBLP:conf/cvpr/ChenZXNSLC17,DBLP:journals/tmm/0007ZXFZ0L19, 8684270, DBLP:conf/mm/LiuZZZW18}, referring expression~\cite{Hu_2016_CVPR,zhao2018weakly,liu2019adaptive}, and visual question answering (VQA)~\cite{Antol_2015_ICCV,DBLP:journals/pami/LiangJCKLH19,yu2019deep, DBLP:journals/tomccap/ZhaLYZ19}. In these works, the co-reference between vision and language is usually performed in a single round. Taking VQA as an example, given an image and a question, the agent identifies the interest areas related to the question and infers an answer. In contrast, visual dialog~\cite{Das_2017_CVPR, Vries_2017_CVPR,DBLP:conf/naacl/KotturMPBR19} is a multi-round extension for VQA. The interactions between the image and multi-round question-answer pairs (history) are progressively changing, and the relationships among the objects in the image are influenced by the current question. Visual dialog is a challenging task due to these underlying semantic dependencies in the textual and visual contexts. Therefore, how to effectively realize the context-aware relational reasoning is vital to the task.

\begin{figure*}[t]
\centering
  \includegraphics[width=17.5cm]{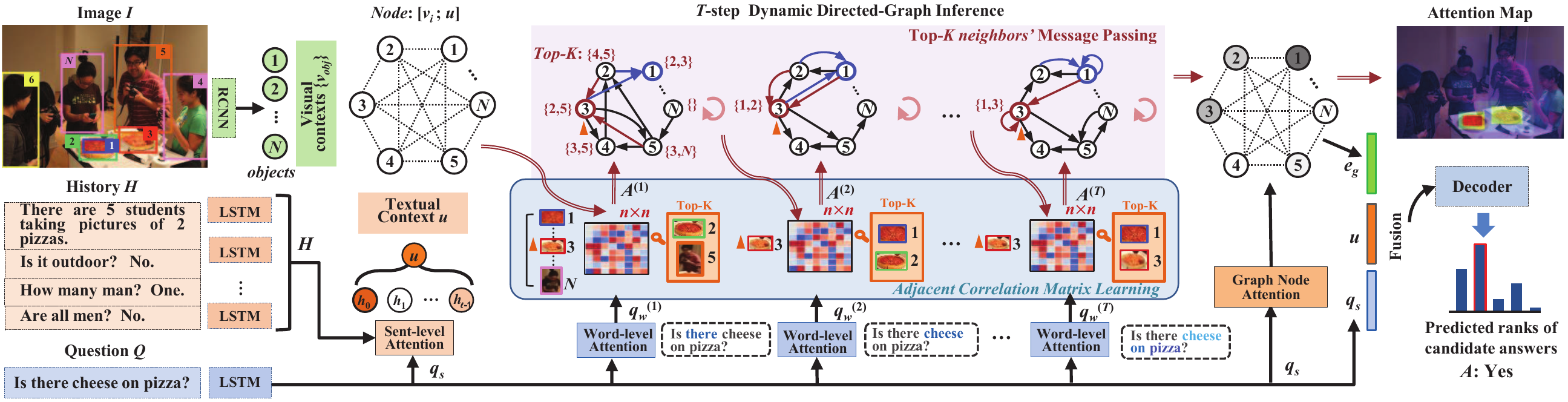}\\
  \caption{The overall framework of Context-Aware Graph. Our context-aware graph is constructed with visual contexts $\{v_{obj}\}$ and textual context $u$. The dynamic relations among the nodes are iteratively inferred via Top-$K$ neighbors' Message Passing under the guidance of word-level question command $q_{w}^{(t)}$. For example, the {\color{red}red} and {\color{blue}blue} nodes in the graph respectively have different top-2 related neighbor nodes, and different directions of the message passing flow on the connected edges.}
  \label{figure2}
\end{figure*}

For relational reasoning, the graph structure is employed to exploit the context-aware co-reference of image and history. Except for the graph structure referring to different multi-modal entities as shown in Fig.~\ref{figure1}, prior graph-based models considered the fixed graph attention or embedding, such as \emph{fixed} fully-connected graph (FGA \cite{schwartz2019factor}), \emph{fixed} once graph attention evolution (FGA \cite{schwartz2019factor}) and \emph{fixed} unidirectional message passing (GNN \cite{zheng2019reasoning}). In this paper, we are inspired by the nature of the visual dialog task, \emph{i.e.}, the dynamic multi-modal co-references in multi-round conversations. Fig.~\ref{figure2} shows the flexibility and adaptivity of our graph-based method, which iteratively evolves by \emph{adaptive top-$K$} and \emph{adaptive-directional} message passing. The significance of our method is that it exploits the image-history co-reference in a dynamic adaptive graph learning mode.

In order to comprehend the complex multi-modal co-reference relationships over time, we propose a Context-Aware Graph (CAG) neural network. As shown in Fig.~\ref{figure1} (c), each node in our graph is a multi-modal context representation, which contains both visual objects and textual history contexts; each edge contains the fine-grained visual interactions of the scene objects in the image. In CAG, all the nodes and edges are iteratively updated through an adaptive top-$K$ message passing mechanism. 
As shown in Fig.~\ref{figure2}, in every message passing step, each graph node adaptively selects the most $K$-relevant nodes, and only receives the messages from them. It means that our CAG solution is an asymmetric dynamic directed graph, which observes adaptive message passing in the graph structure. Note that iterative CAG graph inference is shown as an effective realization of humans' multi-step reasoning~\cite{gan2019multi,DBLP:conf/iclr/HudsonM18}. Finally, after the multi-turn graph inference, we impose a graph attention on all the nodes to obtain the final graph embedding for the answer prediction.

Fig.~\ref{figure2} provides an overview of the proposed CAG. Specifically, CAG consists of three components: (1) \textbf{Graph Construction} (Sec.~\ref{sec:3.1}), which constructs the context-aware graph based on the representations of dialog-history and objects in the image; (2) \textbf{Iterative Dynamic Directed-Graph Inference} (Sec.~\ref{sec:3.2}), the context-aware graph is iteratively updated via $T$-step dynamic directed-graph inference; (3) \textbf{Graph Attention Embedding} (Sec.~\ref{sec:3.3}), which applies a graph attention to aggregate the rich node semantics. Then, we jointly utilize the generated graph, the encoded question, and the history context features to infer the final answer.

The contributions are summarized as follows. We propose a Context-Aware Graph (CAG) neural network for visual dialog, which targets at discovering the partially relevant contexts and building the dynamic graph structure. (1) We build a fine-grained graph representation with various visual objects and attentive history semantics. The context cues on each node not only refer to the joint visual-textual semantic learning, but also involve iterative relational reasoning among image $I$, question $Q$ and history $H$. (2) To eliminate the useless relations among the nodes, we design an adaptive top-$K$ message passing mechanism and a graph attention to pick up more relevant context nodes. Each node has different related neighbors (different relations). As for the same node, the inbound and outbound messages vary from iteration to iteration. (3) Extensive experiments are conducted on VisDial v0.9 and v1.0 datasets, and CAG achieves new state-of-the-art performances out of the pervious graph-based methods.

\section{Relate Work}
\noindent \textbf{Visual Dialog.} For the visual dialog task\cite{Das_2017_CVPR, Vries_2017_CVPR,DBLP:conf/naacl/KotturMPBR19}, 
current encoder-decoder based works can be divided into three facets. (1) \textbf{Fusion-based models.} Late fusion (LF)~\cite{Das_2017_CVPR} and hierarchical recurrent network (HRE)~\cite{Das_2017_CVPR} directly encoded the multi-modal inputs and decoded the answer. (2) \textbf{Attention-based models}. To improve performance, various attention mechanisms have been widely used in the task, including history-conditioned image attention (HCIAE)~\cite{NIPS2017_6635}, sequential co-attention (CoAtt)~\cite{Wu_2018_CVPR}, dual visual attention (DVAN)~\cite{ijcai2019-693}, and recurrent dual attention (ReDAN)~\cite{gan2019multi}. (3) \textbf{Visual co-reference resolution models}. Some attention-based works focused on explicit visual co-reference resolution. Seo et al.~\cite{NIPS2017_6962} designed an attention memory (AMEM) to store previous visual attention distrubution. Kottur et al.~\cite{Kottur_2018_ECCV} utilized neural module networks~\cite{Andreas_2016_CVPR} to handle visual co-reference resolution at word-level. Niu et al.~\cite{niu2019recursive} proposed a recursive visual attention (RvA) mechanism to recursively reviews history to refine visual attention.

\noindent \textbf{Graph Neural Network (GNN).} Graph neural networks have attracted attention in various tasks \cite{wang2019neighbourhood,DBLP:conf/cvpr/Liu0SC18,li2019actional,gu2019scene,8902166}. The core idea is to combine the graphical structural representation with neural networks, which is suitable for reasoning-style tasks. Liu et al.~\cite{DBLP:conf/cvpr/TeneyLH17} proposed the first GNN-based approach for \textbf{VQA}, which built a scene graph of the image and parsed the sentence structure of the question, and calculated their similarity weights. Later, Norcliffe-Brown et al.~\cite{DBLP:conf/nips/Norcliffe-Brown18} modeled a graph representation conditioned on the question, and exploited a novel graph convolution to capture the interactions among different detected object nodes. As for \textbf{visual dialog}, there are merely two related works. Zheng et al.~\cite{zheng2019reasoning} proposed an EM-style GNN to conduct the textual co-reference; it regarded the caption and the previous question-answer (QA) pairs as observed nodes, and the current answer was deemed as an unobserved node inferred using EM algorithm on the textual contexts. Schwartz et al.~\cite{schwartz2019factor} proposed a factor graph attention mechanism, which constructed the graph over all the multi-modal features and estimated their interactions.

Fig.~\ref{figure1} illustrates the difference between our work and other two graph-based models~\cite{schwartz2019factor, zheng2019reasoning}. In this paper, we build a fine-grained context-aware graph, which involves the context co-reference in and between specific objects and history snippets under the guidance of word-level attended question semantics. Apart from the fine-grained object-level features of $I$ (visual contexts) on the node representations, both sentence-level and word-level textual semantics of $H$ and $Q$ are utilized  in the graph.
We implement iterative dynamic message propagation on edges to aggregate the relationships among nodes for answer prediction. In a nutshell, we realize the cross-modal semantic understanding by context-aware relational reasoning.

\section{Our Approach}
The visual dialog task refers to relational learning, which involves complicated semantic dependencies among implicit contexts of image, question and history. How to model the context-aware reasoning is critical.
In this paper, we propose a dynamic directed-graph inference to iteratively review the multi-modal context cues.
Given an image \emph{I} and the dialog history $H\!=\!\lbrace C,\left(q_{1},a_{1} \right),...,\left(q_{\ell-1},a_{\ell-1} \right)\rbrace$, where \emph{C} is the image caption, $(q, a)$ is any question-answer pair and $\ell$ is the turn number of current dialog. The goal of the model is to infer an exact answer for the current question $Q$
by ranking a list of 100 candidate answers $A=\lbrace a_{\ell}^{\left(1\right)},...,a_{\ell}^{\left(100\right)}\rbrace$.
The following sub-sections describe the details of the proposed CAG model.

\subsection{Graph Construction}
\label{sec:3.1}
\noindent \textbf{Feature Representation.} Given an image \emph{I}, we extract the object-level features using Faster-RCNN \cite{Anderson_2018_CVPR} and apply a single-layer MLP with activation \emph{tanh} to encode them into a visual feature sequence $V=\{v_1,...,v_n\} \in \mathbb{R}^{d\times n}$, where
$n$ is the number of detected objects. For the current question $Q$, we first transform it into word embedding vectors $\mathcal{W}^Q=(w_1,...,w_m) \in \mathbb{R}^{d_w\times m}$, where $m$ denotes the number of tokens in $Q$. Then we use an LSTM to encode $\mathcal{W}^Q$ into a sequence $U^Q$ = $(h^q_1,...,h^q_m)\in \mathbb{R}^{d\times m}$, and take the last vector $h^q_m$ as the sentence-level representation of question $Q$, denoted as $q_s=h^q_m$. 
Similarly, we adopt another LSTM to extract the features $U^H$ = $(h_0,...,h_{\ell-1})\in \mathbb{R}^{d \times \ell}$ of history $H$ at sentence-level, where $h_0$ is the embedding feature of image caption \emph{C}.

As questions in a dialog usually have at least one pronoun (\emph{e.g.}, ``it", ``they", ``he"), the dialogue agent is required to discover the relevant textual contexts in the previous history snippets. We employ a question-conditioned attention to aggregate the textual context cues of history, which can be deemed as textual co-reference. The whole process is formulated as follows:
\begin{equation}
\left\{\begin{matrix}
\begin{aligned}
z_h &=tanh((W_{q}q_s)\mathds{1}^\top+W_{h} U^H);\\
\alpha_h &=softmax(P_h z_h);\\
u &= \sum_{j=0}^{\ell-1}\alpha_{h,j} U_{j}^H,
\end{aligned}
\end{matrix}\right.
\label{eq:1}
\end{equation}
where $W_q$, $W_h\in \mathbb{R}^{d \times d}$ and $P_h\in \mathbb{R}^{1 \times d}$ are learnable parameters, $\mathds{1}\in\mathbb{R}^{1 \times \ell}$ is a vector with all elements set to 1, and $\alpha_{h,j}$ and $U_{j}^H$ are respective the $j$-th element of $\alpha_{h}$ and $U^H$. $u\in \mathbb{R}^{d \times 1}$ denotes the history-related textual context and is further used to construct the context-aware graph.

\noindent \textbf{Graph Representation.}
Visual dialog is an on-going conversation. The relations among different objects in the image frequently dynamically vary according to the conversational content. In order to deeply comprehend the conversational content, we build a context-aware graph, which takes both visual and textual contexts into account. The graph structure (relations among objects) will be later iteratively inferred via an adaptive top-$K$ message passing mechanism in Sec.~\ref{sec:3.2}.
Here we construct a graph $G=\{{\cal N},\mathcal{E}\}$, where the $i$-th node ${\cal N}_i$ denotes a joint context feature, corresponding to the $i$-th visual object feature $v_i$ and its related context feature $c_i$; the directed edge $\mathcal{E}_{j\rightarrow i}$ represents the relational dependency from node ${\cal N}_j$ to node ${\cal N}_i$ ($i, j\in [1,n])$. Considering the iterative step $t$, the graph is denoted as $G^{(t)}=\{{\cal N}^{(t)},\mathcal{E}^{(t)}\}$. There are two cases of ${\cal N}^{(t)}$:
\begin{equation}
\left\{\begin{matrix}
\begin{aligned}
{\cal N}^{(t)} &=({\cal N}^{(t)}_1,...,{\cal N}^{(t)}_n);\\
{\cal N}_i^{(t=1)} &=[v_i;u]; \ \ \mathcal{N}_i^{(t>1)} =[v_i;c_i^{(t)}],
\end{aligned}
\end{matrix}\right.
\label{eq:2}
\end{equation}
where $[;]$ is the concatenation operation, textual context $u$ is calculated by Eq.~\ref{eq:1}, and ${\cal N}^{(t)}\in \mathbb{R}^{2d\times n}$. For node ${\cal N}_i^{(t)}$ in the iterative step $t$, the visual feature $v_i$ is fixed, and we focus on the context learning of $c_i^{(t)}$.

\subsection{Iterative Dynamic Directed-Graph Inference}
\label{sec:3.2}
\begin{figure}[t]
\centering
  \includegraphics[width=8cm]{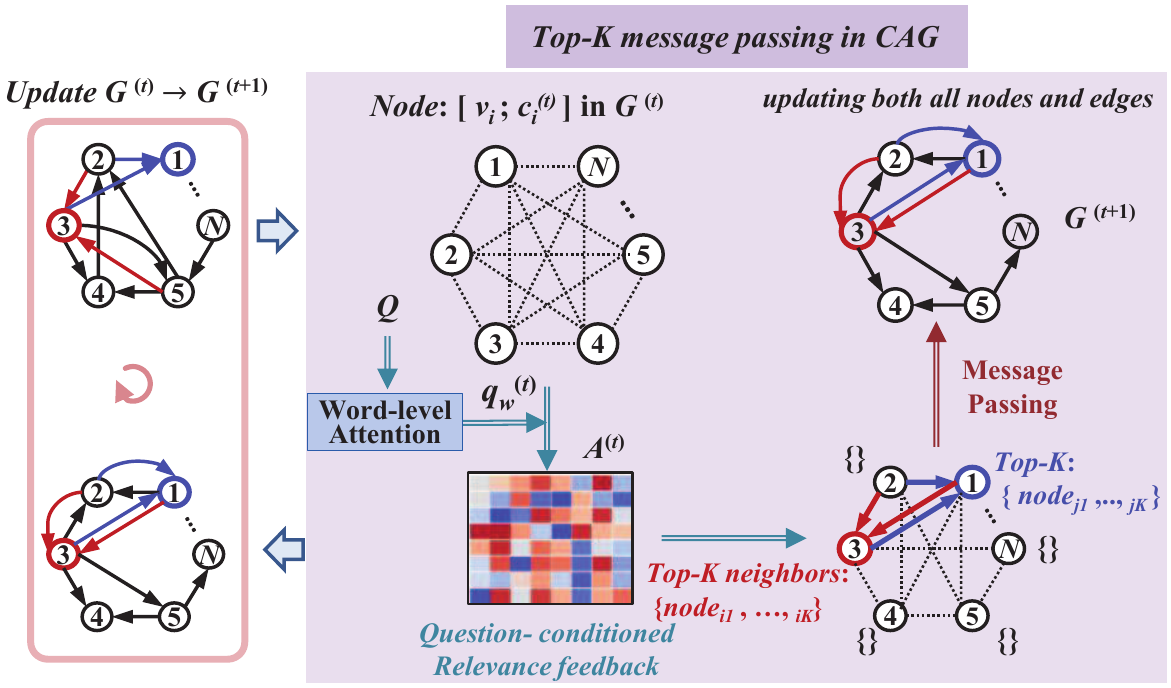}\\
  \caption{Adaptive Top-$K$ Message Passing Unit.
  }\label{figure3}
\end{figure}

Visual dialog contains implicit relationships among the image, question and history. From a technical point of view, the key nature of visual dialog is multi-step reasoning and image-history co-reference. Thus, we address it by adaptively capturing the related visual-historical cues in a dynamic multi-modal co-reference mode. Here, we first exploit attended question commands $\{q_{w}^{(t)}\}$ to instruct message passing on the edges in the graph structure, $t \in [1,T]$, where $T$ is the number of iterations; imitating humans reviewing different keywords multiple times. Then, it is worth noting that in our solution, the ``\textbf{dynamic directed-graph inference}" process, considers flexible, effective, and relevant message propagation. As shown in Fig.~\ref{figure3}, in each iterative step, the context-aware graph is updated through two facets: (1) adjacent correlation learning. Under the instruction of the current question command $q_{w}^{(t)}$, each node adaptively selects the top-$K$ most relevant nodes as its neighbors based on an adjacent correlation matrix; (2) top-$K$ message passing. To capture the dynamic realtions in the  graph, each node receives messages from its top-$K$ neighbors and aggregates these messages to update its context feature.

\noindent \textbf{Question-conditioned Relevance Feedback via Adjacent Correlation Learning.}
To infer a correct answer, we have to discover accurate semantics of question $Q$. In each iterative step $t$, reviewing different words in $Q$ is helpful to locate attentive keywords. Based on the word-level feature sequence of the question $U^Q$ = $(h^q_1,...,h^q_m)$, we employ a self-attention to obtain the word attention distribution $\alpha_{q}^{(t)}$. Then, the word embedding sequence $\mathcal{W}^Q=(w_1,...,w_m)$ is jointly aggregated with $\alpha_{q}^{(t)}$ to get a new attended question feature $q_{w}^{(t)}$. 
\begin{equation}
\left\{\begin{matrix}
\begin{aligned}
z_q^{(t)} &=L2Norm(f_q^{(t)}(U^Q));\\
\alpha_{q}^{(t)} &=softmax(P_q^{(t)} z_q^{(t)});\\
q_{w}^{(t)} &=\sum_{j=1}^{m} \alpha_{q,j}^{(t)} w_{j},
\end{aligned}
\end{matrix}\right.
\label{eq:3}
\end{equation}
where $f_q^{(t)}(.)$ denotes a two-layer MLP and $P_q^{(t)}\in\mathbb{R}^{1\times d}$. The parameters of $f_q^{(t)}(.)$ and $P_q^{(t)}$ are independently learned in the $t$-th iterative step. $q_{w}^{(t)}\in\mathbb{R}^{d_w\times 1}$ is defined as the $t$-th question command at the word level.

After obtaining the question command $q_{w}^{(t)}$, we measure the correlation among different nodes in the graph. We design an adjacency correlation matrix of the graph $G^{(t)}$ as $A^{(t)} \in \mathbb{R}^{n \times n}$, in which each value $A^{(t)}_{l\rightarrow i}$ represents the connection weight of the edge $\mathcal{E}^{(t)}_{l\rightarrow i}$. 
We learn the correlation matrix $A^{(t)}$ by computing the similarity of each pair of two nodes in ${\cal N}^{(t)}$ under the guidance of $q_{w}^{(t)}$.
\begin{equation}
A^{(t)} =(W_{1}{\cal N}^{(t)})^\top ((W_{2}{\cal N}^{(t)})\odot (W_{3} q_{w}^{(t)})),
\label{eq:4}
\end{equation}
where $W_1\in \mathbb{R}^{d\times 2d}$, $W_2\in \mathbb{R}^{d\times 2d}$ and $W_3\in \mathbb{R}^{d\times d_w}$ are learnable parameters, and $\odot$ denotes the hadamard product, $i.e.,$ element-wise multiplication.

A fact is that there are always only a part of the detected objects in the image related to the question (\emph{i.e.}, sparse relationships). Moreover, among these objects, each object is always irrelated with most of other objects. Therefore, each node in the graph is required to connect with the most relevant neighbor nodes. 
In order to learn a set of relevant neighbors $S_i^{(t)}$ of node ${\cal N}_i^{(t)}$, $i\in [1,n]$, $t\in [1,T]$, we adopt a ranking strategy as:
\begin{equation}
S_i^{(t)}=topK(A_i^{(t)}),
\label{eq:5}
\end{equation}
where $topK$ returns the indices of the $K$ largest values of an input vector, and $A_i^{(t)}$ denotes the $i$-th row of the adjacent matrix. Thus, each node has its independent neighbors $S_i^{(t)}$. As shown in Fig.~\ref{figure2} ($topK$, $K$ = $2$), even the same node can have different neighbors in different iterative steps. It means that our solution is an adaptive dynamic graph inference process. Our CAG graph is an asymmetric directed-graph.
\noindent \textbf{Relational Graph Learning via Top-$K$ Message Passing.} The graph structure is now relational-aware. Each node can be affected by its $K$ neighbor nodes. We propagate the relational cues to each node via message passing. Taking node ${\cal N}_i^{(t)}$ in the $t$-th step as an example, it receives the messages from its most $K$ relevant neighbor nodes $\{\mathcal{N}_j^{(t)}\}$, where $j \in S_i^{(t)}$.

To evaluate the influences of relevant neighbors, $B_{j\rightarrow i}^{(t)}$ normalizes the connection weight of edge $\mathcal{E}^{(t)}_{j\rightarrow i}$ (\emph{i.e.}, $\mathcal{N}_j^{(t)}$ $\rightarrow$ $\mathcal{N}_i^{(t)}$). $A^{(t)}_{j\rightarrow i}$ denotes the adjacent correlation weight of edge $\mathcal{E}^{(t)}_{j\rightarrow i}$. As shown in Eq.~\ref{eq:6}, $B_{j\rightarrow i}^{(t)}$ normalizes the weights of the neighbour set $\{ A^{(t)}_{j\rightarrow i}\}$ ($j \in S_i^{(t)}$) with a $softmax$ function. In addition, under the guidance of the question command $q_{w}^{(t)}$, $m_{j\rightarrow i}^{(t)}$ calculates the inbound message of neighbour $\mathcal{N}_j^{(t)}$ to $\mathcal{N}_i^{(t)}$.
At last, ${\cal N}_i^{(t)}$ sums up all the inbound messages to get the final message feature $M_i^{(t)}$. The whole process is formulated as follows:
\begin{equation}
\small
\left\{\begin{matrix}
\begin{aligned}
[B_{j\rightarrow i}^{(t)}] &=\mathop{softmax}\limits_{j \in S_i^{(t)}}([A^{(t)}_{j\rightarrow i}]);\\
m_{j\rightarrow i}^{(t)} &=(W_{4}{\cal N}_j^{(t)}) \odot (W_{5}q_{w}^{(t)});\\
M_i^{(t)} &=\sum \limits_{j \in S_i^{(t)}} B_{j\rightarrow i}^{(t)} m_{j\rightarrow i}^{(t)},
\end{aligned}
\end{matrix}
\right.
\label{eq:6}
\end{equation}
where $W_4 \in \mathbb{R}^{d\times 2d}$ and $W_5 \in \mathbb{R}^{d\times d_w}$ are learnable parameters. $M_i^{(t)}\in \mathbb{R}^{d\times 1}$ denotes the summarized message to $\mathcal{N}_i^{(t)}$, and update $\mathcal{N}_i^{(t)}$ to $\mathcal{N}_i^{(t+1)}$ as follows:
\begin{equation}
\left\{\begin{matrix}
\begin{aligned}
c_i^{(t+1)} &=W_6[c_i^{(t)};M_i^{(t)}];\\
\mathcal{N}_i^{(t+1)} &=[v_i;c_i^{(t+1)}],
\end{aligned}
\end{matrix}\right.
\label{eq:7}
\end{equation}
where $W_6 \in \mathbb{R}^{d\times 2d}$ is a learnable parameter.  Parameters $W_{1} \sim W_{6}$ in Eqs.~\ref{eq:4} $\sim$ \ref{eq:7} are shared for each iteration. After performing $T$-step message passing iterations, the final node representation is denoted as $\mathcal{N}^ {(T+1)}$.
\subsection{Graph Attention Embedding}
\label{sec:3.3}
Up to now, the context learning on each node in the graph $\mathcal{N}^ {(T+1)}$ not only integrates original visual and textual features, but also involves iterative context-aware relational learning. As the majority of questions merely pay attention to a small part of objects in the image scene, we apply a question-conditioned graph attention mechanism to attend all the nodes. The graph attention is learned as follows:
\begin{equation}
\left\{\begin{matrix}
\begin{aligned}
z_g &=tanh((W_{g_1}q_s)\mathds{1}^\top+W_{g_2} \mathcal{N}^{(T+1)});\\
\alpha_g &=softmax(P_g z_g);\\
e_g &= \sum_{j=1}^{n}\alpha_{g,j} \mathcal{N}_j^{(T+1)},
\end{aligned}
\end{matrix}\right.
\label{eq:8}
\end{equation}
where $W_{g1} \in \mathbb{R}^{d\times d}$ and $W_{g2} \in \mathbb{R}^{d\times 2d}$ are learnable parameters. $e_g\in \mathbb{R}^{2d\times 1}$ denotes the attended graph embedding.

Finally, we fuse it with the history-related textual context $u$ and the sentence-level question feature $q_s$ to output the multi-modal embedding $\widetilde{e}$:
\begin{equation}
\widetilde{e} =tanh(W_e [e_g ;u;q_s]).
\label{eq:9}
\end{equation}
The output embedding $\widetilde{e}$ is then fed into the discriminative decoder~\cite{NIPS2017_6635} to 
choose the answer with the highest probability as the final prediction. The details of the training settings are explained in Sec.~\ref{sec:4.2}.

\section{Experiments}
\label{sec:exps}
\subsection{Experiment Setup}
\noindent \textbf{Datasets.} Experiments are conducted on benchmarks VisDial v0.9 and v1.0 ~\cite{Das_2017_CVPR}. VisDial v0.9 contains 83k and 40k dialogs on COCO-train and COCO-val images~\cite{10.1007/978-3-319-10602-1_48} respectively, totally 1.2M QA pairs. VisDial v1.0 is an extension of VisDial v0.9, which adds additional 10k dialogs on Flickr images. The new train, validation, and test splits contains 123k, 2k and 8k dialogs, respectively. Each dialog in VisDial v0.9 consists of 10-round QA pairs for each image. In the test split of VisDial v1.0, each dialog has flexible $m$ rounds of QA pairs, where $m$ is in the range of 1 to 10.

\noindent \textbf{Implementation Details.}
\label{sec:4.2}
The proposed method is implemented on the platform of Pytorch. We build the vocabulary that contains the words occurring at least 4 times in the training split. And the captions, questions, and answers are truncated to 40, 20 and 20, respectively. Each word in the dialog is embedded into a 300-dim vector by the GloVe embedding initialization~\cite{pennington2014glove}. We adopt Adam optimizer~\cite{DBLP:journals/corr/KingmaB14} and initialize the learning rate with $4\times 10^{-4}$. The learning rate is multiplied by 0.5 after every 10 epochs. We set all the LSTMs in the model with 1-layer and 512 hidden states, and apply Dropout~\cite{srivastava2014dropout} with ratio 0.3 for attention layers and the last fusion layer. Finally, the model is trained with a multi-class $N$-pair loss~\cite{guo2019image,NIPS2017_6635}.
\subsection{Ablation Study of CAG}
We evaluate two main hyperparameters in our model CAG -- the selected neighbor number $K$ and the number of iterative steps $T$, and validate the influence of visual features and main components of CAG.

\begin{figure}[t]
\setlength{\abovecaptionskip}{0.05cm}
\centering
  \includegraphics[width=8.2cm]{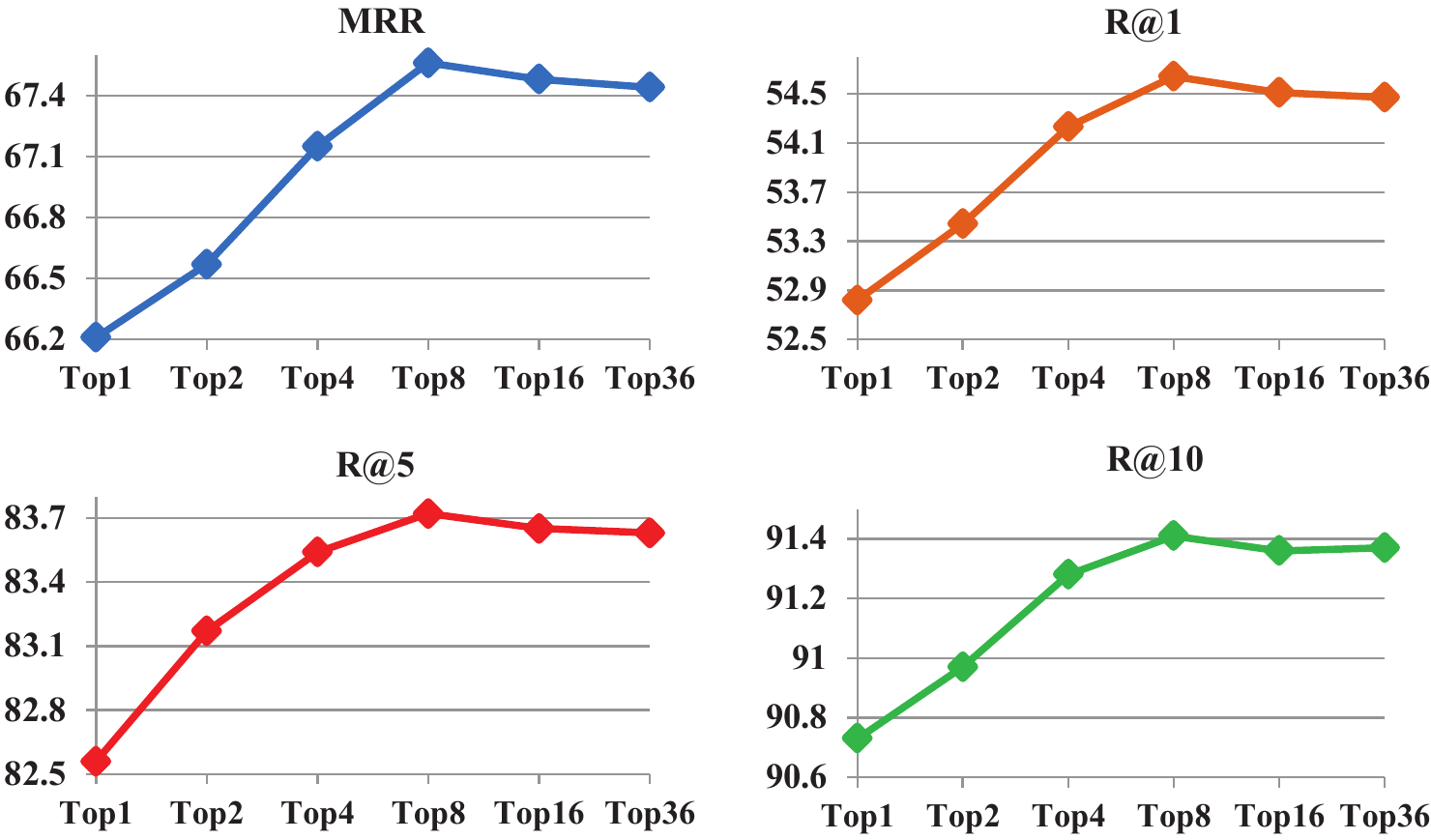}\\
  \caption{Performance comparison of the neighbor number $K$ on VisDial val v0.9
  }
  \label{figure4}
\end{figure}

\noindent \textbf{Neighbor Number $(K)$.} We test different neighbor numbers $K\in \{1, 2, 4, 8, 16, 36\}$. As shown in Fig.~\ref{figure4}, $K=8$ is an optimal parameter setting. Performances drop significantly for $K<8$. It means that if the selected neighbor nodes are insufficient, the relational messages can not be fully propagated. While setting the neighbor number $K>8$, the node receiving redundant irrelevant messages from neighbors can disturb the reasoning ability of the model. Thus, we set the neighbor number $K=8$ in the following experiments.
\label{sec:4.3}

\noindent \textbf{Iterative step $(T)$.}  $T$ indicates the number of relational reasoning steps to arrive the answer. We test different steps $T$ to analysis the influences of iterative inferences. As shown in Table~\ref{table1}, the performance of CAG is gradually improved with the increasing $T$. We have the best performance while $T=3$, lifting R@1 from 53.25 ($T$ = 1) to 54.64. The proposed iterative graph inference is effective. Visualization results in Fig.~\ref{figure5} further validate this result. When $T>3$, the performance drops slightly. It means that if the relations in the graph have been fully inferred, further inference does not help. The questions in the VisDial datasets are collected from relative simple free-form human dialogue. The setting of $T$ = 3 already performs well. In the following experiment, we set $T=3$.

\noindent \textbf{Main Component Comparison.} A few variants are proposed for ablation study. \textbf{CAG w/o Infer} denotes that CAG removes the whole dynamic directed-graph inference in Sec.~\ref{sec:3.2}. It means that the relations in the graph including all the nodes and edges will not be updated and inferred. \textbf{CAG w/o \emph{u}} denotes that CAG without textual-history context $u$, where the whole graph merely describes visual context cues. \textbf{CAG w/o Q-att} denotes CAG without word-level attention on question $Q$. 
\textbf{CAG w/o G-att} removes the graph attention module, where all the node representations are average pooled to get the final graph embedding.

\begin{table}
\centering
\fontsize{9.5}{10}\selectfont
\resizebox{0.48\textwidth}{!}{
\begin{tabular}{lcccccc}
\hline
Model  &Step $T$ & Mean$\downarrow$ & MRR$\uparrow$ & R@1$\uparrow$ & R@5$\uparrow$ & R@10$\uparrow$ \\
\hline
\multirow{4}*{CAG}
    & $T$ = 1 & 4.02 & 66.32  & 53.25 & 82.54   & 90.55  \\
    & $T$ = 2 & 3.91 & 66.93  & 53.76 & 83.11   & 90.96 \\
    & $T$ = 3 &\textbf{3.75} &\textbf{67.56} &\textbf{54.64} &\textbf{83.72} &\textbf{91.48}\\
    & $T$ = 4 & 3.83 & 67.28  & 54.11 & 83.46   & 91.17 \\
\hline
CAG w/o Infer   & -    & 4.11 & 65.73  & 52.56 & 82.38 & 90.36 \\
CAG w/o \emph{u}   & $T$ = 3  & 4.19 & 65.26  & 51.83 & 81.55   & 90.21  \\
CAG w/o Q-att   & $T$ = 3  & 3.91 & 66.70  & 53.74 & 82.75   & 90.89  \\
CAG w/o G-att   & $T$ = 3 & 3.86 & 66.98  & 53.99 & 83.08   & 91.04 \\
CAG       & $T$ = 3 &\textbf{3.75} &\textbf{67.56} &\textbf{54.64} &\textbf{83.72} &\textbf{91.48}\\
\hline
\end{tabular}}
\caption{Ablation studies of different iterative steps $T$ and the main components on VisDial val v0.9.}
\label{table1}
\end{table}

As shown in Table~\ref{table1}, compared with \textbf{CAG}, \textbf{CAG w/o Infer} drops MRR significantly from 67.56 to 65.73. It indicates that the graph inference effectively performs well for relational reasoning. Learning the implicit relations among the nodes is helpful to predict the final answer. \textbf{CAG w/o \emph{u}} drops R@1 significantly from 54.64 to 51.83. It indicates that the joint visual-textual context learning is necessary. Relations among the nodes can not be fully inferred without textual cues $u$. \textbf{CAG w/o Q-att}, which replaces question commands $\{ q_{w}^{(t)} \}$ with the sentence-level feature $q_s$, drops R@1 from 54.64 to 53.74. It also can be explained. Figs.~\ref{figure5}$\sim$\ref{figure7} demonstrate that the attentive words always vary during the inference process. It usually firstly identifies the target in the question, than focuses on related objects, and finally observes attributes and relations in both visual and textual context cues to infer the answer. \textbf{CAG w/o G-att}, which removes the final graph attention module, drops R@1 from 54.64 to 53.99. Although each node involves relational reasoning, not all the nodes are relevant to the current question. Therefore, paying attention to relevant nodes in the relational graph is helpful to infer an exact answer.
\begin{table}
\centering
\fontsize{9.5}{10}\selectfont
\resizebox{0.48\textwidth}{!}{
\begin{tabular}{lccccc}
\hline
Model  & Mean$\downarrow$ & MRR$\uparrow$ & R@1$\uparrow$ & R@5$\uparrow$ & R@10$\uparrow$ \\
\hline
\multicolumn{6}{c}{Attention-based Models}\\
\hline
  HCIAE~\cite{NIPS2017_6635}       &4.81 &62.22 &48.48 &78.75 &87.59 \\
  AMEM~\cite{NIPS2017_6962}        &4.86 &62.27 &48.53 &78.66 &87.43\\
  CoAtt~\cite{Wu_2018_CVPR}        &4.47 &63.98 &50.29 &80.71 &88.81\\
  DVAN-VGG~\cite{ijcai2019-693} &4.38 &63.81 &50.09 &80.58 &89.03\\
  RvA-VGG~\cite{niu2019recursive} &4.22 &64.36 &50.40 &81.36 &89.59\\
  HACAN-VGG~\cite{yang2019making}       &4.32 &64.51 &50.72 &81.18 &89.23\\
\hline
\multicolumn{6}{c}{Graph-based Models}\\
\hline
  GNN~\cite{zheng2019reasoning}    &4.57 & 62.85  & 48.95 & 79.65 &88.36 \\
  FGA w/o Ans~\cite{schwartz2019factor} &4.63 &62.94 &49.35  & 79.31 &88.10\\
  CAG-VGG (Ours) & \textbf{4.13} & \textbf{64.91}  & \textbf{51.45} & \textbf{81.60} & \textbf{90.02} \\
  CAG (Ours)       &\textbf{3.75} &\textbf{67.56} &\textbf{54.64} &\textbf{83.72} &\textbf{91.48} \\
\hline
\end{tabular}}
\caption{Performance comparison on VisDial val v0.9 with VGG features. Our model with VGG features is denoted as CAG-VGG.}
\label{table2}
\end{table}

\begin{table*}[t]
  \centering
  \fontsize{9}{9.5}\selectfont
  \begin{threeparttable}
    \begin{tabular}{lccccc|cccccc}
    \toprule
    \multirow{2}{*}{Model}&
    \multicolumn{5}{c}{ VisDial v0.9 (val)}&\multicolumn{6}{c}{ VisDial v1.0 (test-std)}\cr
    \cmidrule(lr){2-6} \cmidrule(lr){7-12}
    &Mean$\downarrow$ &MRR$\uparrow$ &R@1$\uparrow$ &R@5$\uparrow$ &R@10$\uparrow$  &Mean$\downarrow$ &NDCG$\uparrow$ &MRR$\uparrow$ &R@1$\uparrow$ &R@5$\uparrow$ &R@10$\uparrow$\cr
    \midrule
     &\multicolumn{10}{c}{Fusion-based Models}\\
    \cline{1-12}
    LF~\cite{Das_2017_CVPR}          &5.78   &58.07 &43.82 &74.68 &84.07    &5.95   &45.31 &55.42 &40.95 &72.45 &82.83 \cr
    HRE~\cite{Das_2017_CVPR}         &5.72   &58.46 &44.67 &74.50 &84.22    & 6.41 &  45.46 & 54.16  & 39.93 & 70.45 & 81.50 \cr
    \hline
    &\multicolumn{10}{c}{Attention-based Models}\\
    \cline{1-12}
    HREA~\cite{Das_2017_CVPR}        &5.66 &58.68 &44.82 &74.81 &84.36     &- &- &- &- &- &-\cr
    MN~\cite{Das_2017_CVPR}          &5.46 &59.65 &45.55 &76.22 &85.37     &5.92 &47.50 &55.49 &40.98 &72.30 &83.30\cr
    HCIAE~\cite{NIPS2017_6635}       &4.81 &62.22 &48.48 &78.75 &87.59     &- &- &- &- &- &-\cr
    AMEM~\cite{NIPS2017_6962}        &4.86 &62.27 &48.53 &78.66 &87.43     &- &- &- &- &- &-\cr
    CoAtt~\cite{Wu_2018_CVPR}        &4.47 &63.98 &50.29 &80.71 &88.81     &- &- &- &- &- &-\cr
    CorefNMN~\cite{Kottur_2018_ECCV} &4.45 &64.10 &50.92 &80.18 &88.81     &4.40 &54.70 &61.50 &47.55 &78.10 &88.80\cr
    DVAN~\cite{ijcai2019-693}        &3.93 &66.67 &53.62 &82.85 &90.72     &4.36 &54.70 &62.58 &48.90 &79.35 &89.03\cr
    RVA~\cite{niu2019recursive}      &3.93 &66.34 &52.71 &82.97 &90.73     &4.18 &55.59 &63.03 &49.03 &80.40 &89.83\cr
    Synergistic~\cite{guo2019image}  &- &- &- &- &-                        &4.17 &57.32 &62.20 &47.90 &80.43 &89.95\cr
    DAN~\cite{kang2019dual}          &4.04 &66.38 &53.33 &82.42 &90.38     &4.30 &\textbf{57.59} &63.20 &49.63 &79.75 &89.35\cr
    HACAN~\cite{yang2019making}      &3.97 &\textbf{67.92} &\textbf{54.76} &83.03 &90.68     &4.20 &57.17 &\textbf{64.22} &\textbf{50.88} &80.63 &89.45\cr
    \hline
    &\multicolumn{10}{c}{Graph-based Models}\\
    \cline{1-12}
    GNN~\cite{zheng2019reasoning}    &4.57 &62.85 &48.95 &79.65 &88.36     &4.57 &52.82 &61.37 &47.33 &77.98 &87.83\cr
    FGA w/o Ans~\cite{schwartz2019factor}    &4.63 &62.94 &49.35  & 79.31 &88.10     &- &- &- &- &- &-\cr
    FGA~\cite{schwartz2019factor}    &4.35 &65.25 &51.43 &82.08 &89.56     &4.51 &52.10 &\underline{63.70} &49.58 &\textbf{\underline{80.97}} &88.55 \cr
    CAG (Ours)                  &\textbf{\underline{3.75}} &\underline{67.56} &\underline{54.64} &\textbf{\underline{83.72}} &\textbf{\underline{91.48}}    &\textbf{\underline{4.11}} &\underline{56.64} &63.49 &\underline{49.85} &80.63 &\textbf{\underline{90.15}}\cr
    \bottomrule
    \end{tabular}
    \end{threeparttable}
\caption{Main comparisons on both VisDial v0.9 and v1.0 datasets using the discriminative decoder~\cite{NIPS2017_6635}.}
\label{table3}
\end{table*}

\noindent \textbf{Test with VGG Features.} As some existing methods evaluated with VGG features, to be fair, we test our model with VGG features too. Table 2 shows that our \textbf{CAG-VGG} still outperforms the previous methods that only utilize VGG features. Compared to \textbf{CAG-VGG}, \textbf{CAG} gets a significant performance boost. It indicates the object-region features provide richer visual semantics than VGG features.

\subsection{Comparison Results}
\label{sec:4.3}
\noindent \textbf{Baseline methods.}  In our experiment, compared methods can be grouped into three types: (1) \textbf{Fusion-based Models} (LF~\cite{Das_2017_CVPR} and HRE~\cite{Das_2017_CVPR}); (2) \textbf{Attention-based Models} (HREA~\cite{Das_2017_CVPR}, MN~\cite{Das_2017_CVPR}, HCIAE~\cite{NIPS2017_6635}, AMEM~\cite{NIPS2017_6962}, CoAtt~\cite{Wu_2018_CVPR}, CorefNMN~\cite{Kottur_2018_ECCV}, DVAN~\cite{ijcai2019-693}, RVA~\cite{niu2019recursive}, Synergistic~\cite{guo2019image}, DAN~\cite{kang2019dual}, and HACAN~\cite{yang2019making}); and (3) \textbf{Graph-based Methods} (GNN~\cite{zheng2019reasoning} and FGA~\cite{schwartz2019factor}).

\noindent \textbf{Results on VisDial v0.9.} As shown in Table 3, CAG consistently outperforms most of methods. Compared with fusion-based models \textbf{LF}~\cite{Das_2017_CVPR} and \textbf{HRE}~\cite{Das_2017_CVPR}, the R@1 performance of our \textbf{CAG} is significantly improved, lifting each other by 10.8\% and 9.9\%. For attention-based models, compared to 
\textbf{DAN}~\cite{kang2019dual}, \textbf{CAG} outperforms it at all evaluation metrics. 
\textbf{HACAN}~\cite{yang2019making} reports the recent best results. It first pre-trained with $N$-pair loss, and then used the wrong answers to ``tamper'' the truth-history for data augment. Finally, the truth- and fake-history were used to fine-tune its model via reinforcement learning. Without the fine-tuning tactic, our model \textbf{CAG} still outperforms \textbf{HACAN} on Mean, R@5, and R@10. 

Here, we mainly compare our method with the graph-based models. \textbf{GNN}~\cite{zheng2019reasoning} is a recently proposed method, which constructs a graph exploring the dependencies among the textual-history. In contrast, our \textbf{CAG} builds a graph over both visual-objects and textual-history contexts. 
Compared with \textbf{GNN}, our model achieves 5.7\% improvements on the R@1 metric. \textbf{FGA}~\cite{schwartz2019factor} is the state-of-the-art graph-based method for visual dialog, which treats the candidate answer embedding feature $A$ as new context cue and introduces it into the multi-modal encoding training. This operation improves their results a lot (FGA w/o Ans $vs.$ FGA). Without candidate answer embedding, our model still performs better results, lifting R@1 from 51.43 to 54.64, and decreasing the Mean from 4.35 to 3.75. These comparisons indicate that in our solution, the fine-grained visual-textual joint semantics are helpful for answer inferring.

\noindent \textbf{Results on VisDial v1.0.}
A new metric NDCG (Normalized Discounted Cumulative Gain)~\cite{DBLP:conf/colt/WangWLHL13} is proposed to evaluate quantitative semantics, which penalizes low ranking correct answers. Other metrics accord to evaluate the rank of the ground-truth in the candidate answer list. NDCG tackles more than one plausible answers in the answer set. Compared with the attention-based models, as above mentioned, \textbf{HACAN}~\cite{yang2019making} trained the model twice and \textbf{Synergistic}~\cite{guo2019image} sorted the candidate answers twice. 
Without resorting or fine-tuning, under end-to-end training, our model still performs better performance on the Mean value. 
Compared with the graph-based models, our model has greatly improved the NDCG value. CAG outperforms \textbf{GNN}~\cite{zheng2019reasoning} and \textbf{FGA}~\cite{schwartz2019factor} by 3.8\% and 4.5\% respectively. This also proves that our graph can infer more plausible answers. In addition, we give more intuitive visualization results of CAG to explain the reasoning process detailed in Sec. \ref{sec:4.4}.

\subsection{Qualitative Results}
\label{sec:4.4}
\begin{figure*}[t]
\centering
  \includegraphics[width=17.4cm]{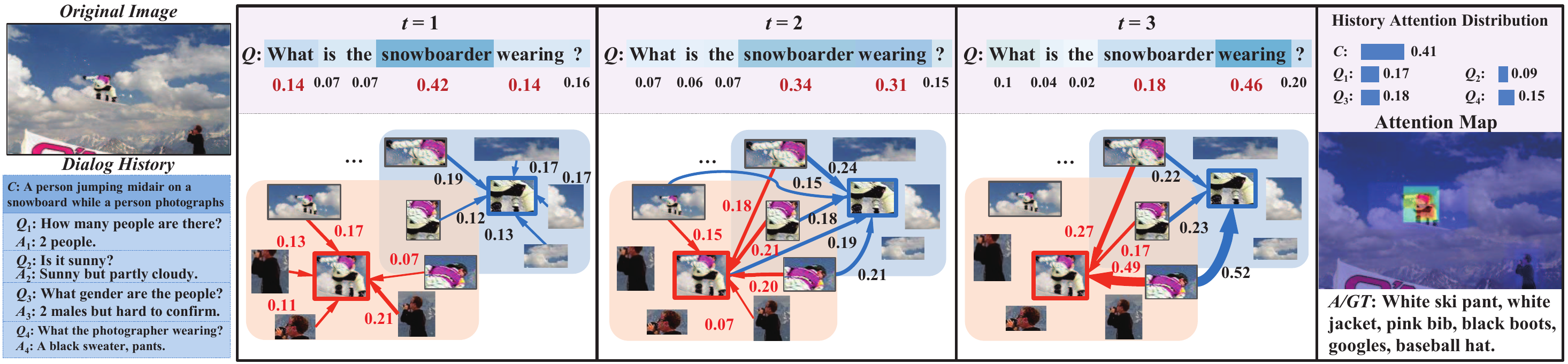}\\
  \caption{Visualization results of iterative context-aware graph inference. It shows the word-level attention on question $Q$, and dynamic graph inference of the top-2 attended objects ({\color{red}red} and {\color{blue}blue} bounding boxes) in image $I$. The number on each edge denotes the normalized connection weight, displaying the message influence propagated from neighbors. There are some abbreviations as follows: question ($Q$), generated answer ($A$), caption ($C$) and the ground-truth ($GT$).
  }
  \label{figure5}
\end{figure*}

\begin{figure*}
\centering
  \includegraphics[width=17.2cm]{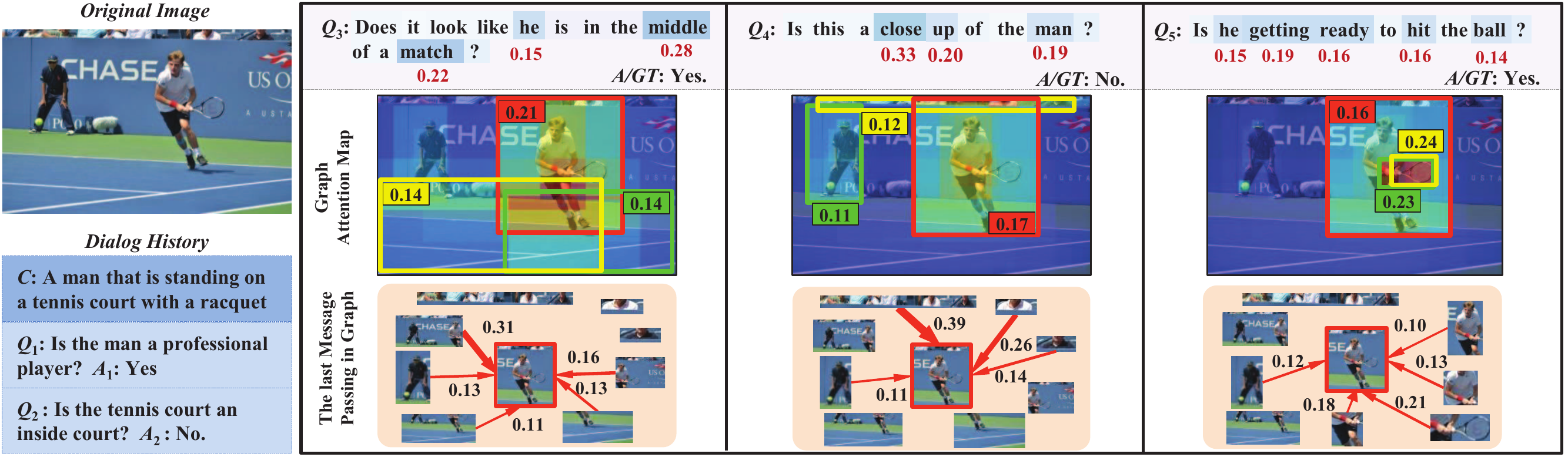}\\
  \caption{Visualization result of a progressive multi-round dialog inference. Each column displays the graph attention map overlaying the image $I$ and the last step of message passing process of the most salient object. In these graph attention maps, bounding boxes correspond to the top-3 attended object nodes, and the numbers along with the bounding boxes represent the node attention weights.
  }
  \label{figure6}
  \vspace{-0.5ex}
\end{figure*}

To further demonstrate the interpretability of our solution, we show an iterative graph inference example in Fig.~\ref{figure5}. Two most salient objects (``snowboarder" and ``pant") in the graph attention maps are selected to display the inference processes. In iterative step $t=1$, the question focuses on the word ``snowboarder" (target). By reviewing the dialog context, ``snowboarder" is related with ``midair" and ``photographer". The above two salient objects receive messages from their relevant neighbor object nodes. Then in iterative step $t=2$, the question changes the attention to both words of ``snowboarder" and ``wearing" (related objects). These two object nodes dynamically update their neighbor nodes under the guidance of the current question command $q_w^{(t=2)}$. In the last step $t=3$, question $Q$ focuses on the word ``wearing" (relation). The edge connections in the graph are further refined by receiving messages from wearing-related nodes. Through multi-step message passing, our context-aware graph progressively finds out much more implicit question-related visual and textual semantics. Finally, the graph attention map overlaying the image $I$ also demonstrates the effectiveness of the graph inference.

We provide another example in Fig.~\ref{figure6} to display relational reasoning in multi-round QA pairs. The edge relations and the nodes' attention weights dynamically vary corresponding to the current question. Our context-aware graph effectively models this dynamic inference process via adaptive top-$K$ message passing. Each node only receives strong messages from the most relevant nodes. The graph attention maps overlaying the image at different rounds further validate the adaptability of our graph on relational reasoning.

In addition, we display the visualization of attentive word clouds on VisDial v1.0. Fig.~\ref{figure7} describes the word-level attention distribution of question $Q$ in different iterative steps. In iterative step $t=1$, the proposed CAG inclines to pronouns in the questions, \emph{e.g.,} ``there", ``it", ``you", ``they". CAG tries to tackle the textual co-reference in the initial relational reasoning. Then, in step $t=2$, CAG prefers to attend nouns related to target objects or associated objects in the image, \emph{e.g.,} ``people", ``building", ``tree". This means CAG trends to infer the relationships among different related objects, namely visual-reference. In the time step $t=3$, the model considers the words that describe the attributes or relations of the objects, \emph{e.g.,} ``color", ``wearing", ``other", ``on". All these appearances indicate that we reasonably and actively promote the iterative inference process using the context-aware graph CAG.

\section{Conclusion}
\begin{figure}[t]
\centering
  \includegraphics[width=8.2cm]{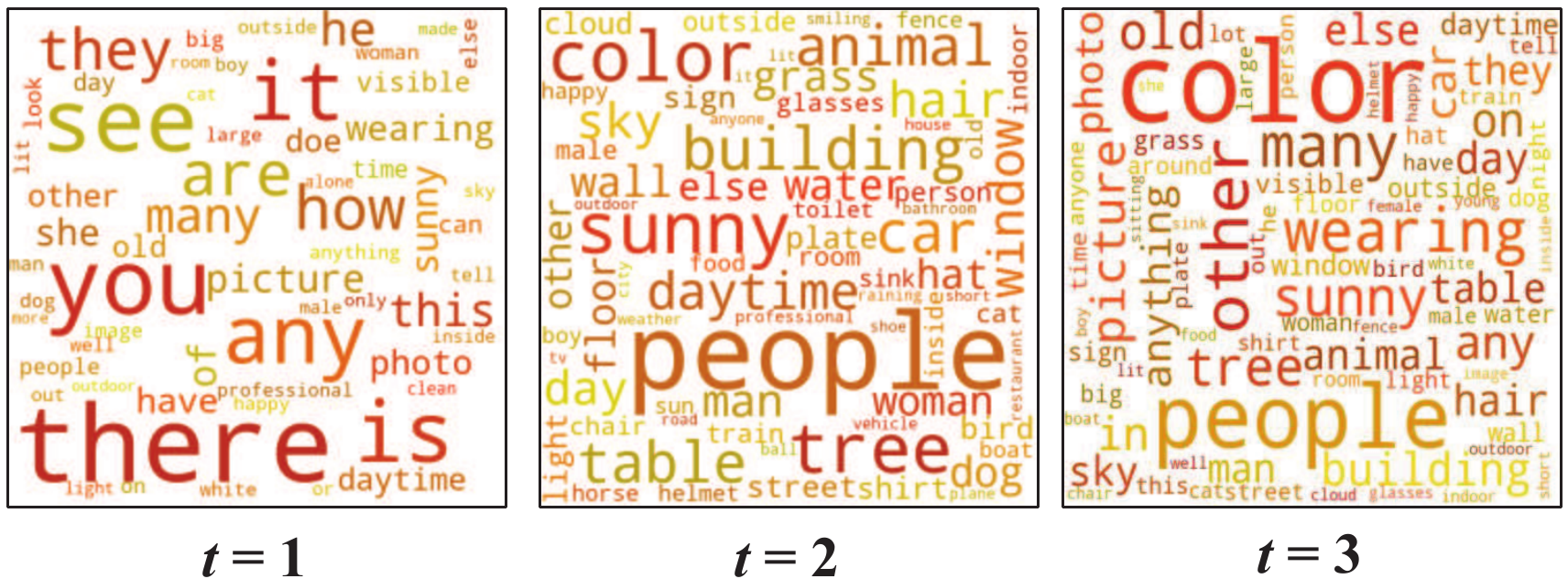}\\
  \caption{Visualization of attentive word cloud of all the questions $\{Q\}$ in different iterative steps on VisDial v1.0.
  }
  \label{figure7}
  \vspace{-1ex}
\end{figure}
In this paper, we propose a fine-grained Context-Aware Graph (CAG) neural network for visual dialog, which contains both visual-objects and textual-history context semantics. An adaptive top-$K$ message passing mechanism is proposed to iteratively explore the context-aware representations of nodes and update the edge relationships for a better answer inferring. Our solution is a dynamic directed-graph inference process. Experimental results on the VisDial v0.9 and v1.0 datasets validate the effectiveness of the proposed approach and display explainable visualization results.
\section*{Acknowledgements}
This work is supported by the National Natural Science Foundation of China (NSFC) under Grants 61725203, 61876058, 61732008, 61622211, and U19B2038.

{\small
\bibliographystyle{ieee_fullname}
\bibliography{egbib}
}

\newpage
\clearpage
\appendix
\section{Supplementary Material}
This supplementary document is organized as follows:
\begin{itemize}
\item We explain two different textual attention mechanisms in Sec. \ref{sec:s1}, especially for the word-level attention on question $Q$ at different iterative processes.
 \item Sec. \ref{sec:s2} elaborates the motivation of the fine-grained graph construction and the relational measurement between different nodes.
\item In the paper, Figs.~\ref{figure5}$\sim$\ref{figure7} illustrate the visualization results on image $I$ and question $Q$. Here, Sec. \ref{sec:s3} mainly demonstrates the influence of history-related context $u$ in Fig.~\ref{figure9}, and supplements additional qualitative results of visual reference in Fig.~\ref{figure10}.
\end{itemize}

\subsection{Textual Attention Mechanisms}
\label{sec:s1}
\begin{itemize}
\item \textbf{Sentence-level attention on history $H$}: The latent attention variable $z_h $ in Eq.~\ref{eq:1} tackles the textual co-reference between $q_s$ and $U^H$, where both $q_s$ and $U^H$ are sentence-level semantics.
\item \textbf{Word-level attention on question $Q$}: For computing word-level attention on $Q$, the latent attention variable $z_q^{(t)}$ in Eq.~\ref{eq:3} measures the question itself. We unfold the MLP operation in Eq.~\ref{eq:3} as follows:
\begin{equation}
\left\{\begin{split}
f_q^{(t)} & = tanh(W_{f_1}U^Q) \odot \sigma(W_{f_2}U^Q);\\
z_q^{(t)} & = L2Norm(f_q^{(t)}(U^Q)),
\end{split}\right.\tag{A1}
\end{equation}
where $W_{f_1}$, $W_{f_2} \in \mathbb{R}^{d\times d}$ are learnable parameters.
\end{itemize}

The operation $f_q^{(t)}$ uses the tangent \& sigmoid activation gates to learn a new word-level feature sequence of question $Q$. Then, the $L2Norm$ operation normalizes each word's new feature embedding vector on the feature dimension. 
With the L2 normalization, $z_q^{(t)} \in \mathbb{R}^{d\times m}$ can equitably evaluate each word in the word sequence of $Q$. Figs.~\ref{figure5}$\sim$\ref{figure7} (especially Fig.~\ref{figure7} validate the adaptability of the word-level attention in different iterative steps.

\subsection{Fine-grained Graph Construction}
\label{sec:s2}
\subsubsection{Node components: visual and textual contexts}
\textbf{One argue maybe that} why we realize the graph initialization with history-related context $u$ and implement the joint visual-textual context learning? The motivation is that without history-related context $u$, the dialog agent can't understand the previous dialogue topic well, nor it can further solve the current visual reference well.

Technically, one challenge of visual dialog is to explore the latent relations among {\color{red}{\textbf{image}}}, {\color{blue}{\textbf{history}}} and {\color{cyan}{\textbf{question}}}. We reformulate the idea of the dynamic graph learning in the paper as follows. In iterative step $t$, as the definition of node ${\cal N}_i^{(t)}$ = $[v_i;c_i^{(t)}]$, $v_i$ denotes the visual feature of object $obj_i$, and $c_i^{(t)}$ records the relevant context related to $obj_i$. $c_i$ considers both $\{v_i\}$ and $u$, and is guided by the question command $q_{w}^{(t)}$.
\begin{equation}
\left\{\begin{split}
\mathcal{N}_i^{(1)} &=[v_i;c_i^{(1)}]=[{\color{red}{\bm{v_i}}};{\color{blue}{\bm{u}}} ];\\
c_i^{(t+1)} &= \emph{\textbf{MP}}\Big(\!  {\color{cyan}{\bm{q_{w}^{(t)}}}} \! \odot\!  \{{\cal N}_j^{(t)}\}|_{top-K}\!  \succ\!  {\cal N}_i^{(t)}\Big)\!  \bowtie\!  c_i^{(t)};\\
\mathcal{N}_i^{(t+1)} &=[v_i;c_i^{(t+1)}],\quad i \in [1,n], t\in [1,T],
\end{split}\right.\tag{A2}
\end{equation}
where $\mathcal{N}_i^{(1)}$ denotes the graph initialization, $\emph{\textbf{MP}}$ denotes the message passing calculation by Eq.~\ref{eq:6}, $\succ$ denotes the adjacent correlation matrix learning by Eq.~\ref{eq:4}, and $\bowtie$ means that the context $c_i^{(t)}$ is updated by Eq.~\ref{eq:7}.

The joint context learning of $c_i^{(t)}$ involving $u$ plays an important role in the graph inference. Both ablation studies in Table~\ref{table1} and qualitative results in Sec.~\ref{sec:s3} detailed below demonstrate the effectiveness. In addition, Fig.~\ref{figure7} also verifies the significance of $u$ in step $t$=1. The introduce of $u$ is helpful to tackle the visual-textual co-reference related to the question, such as parsing pronouns in the question ($e.g.,$ ``he", ``it" and ``there") and grounding the relevant objects in the image.

\subsubsection{Adjacent correlation matrix learning}
\textbf{Another argue maybe that} why impose the question command $q_w^{(t)}$ on only one node side of the matrix $A^{(t)}$ in Eq.~\ref{eq:4}, which is not a symmetrical operation as mutual correlation calculation. We define a classical mutual (symmetrical) correlation calculation as \textbf{CAG-DualQ} as follows:
\begin{equation}
\left\{\begin{split}
 &\textbf{CAG}:\\
 &A^{(t)} =(W_{1}{\cal N}^{(t)})^\top \Big [(W_{2}{\cal N}^{(t)})\! \odot\!  (W_{3} q_{w}^{(t)})\Big ];\\
 &\textbf{CAG-DualQ}:\\
 &A^{(t)} =\Big [(W_{1}{\cal N}^{(t)})\! \odot\!  (W_{3}^{'} q_{w}^{(t)})\Big ]^\top \Big [(W_{2}{\cal N}^{(t)})\! \odot\!  (W_{3} q_{w}^{(t)})\Big ].
\label{eq:12}
\end{split}\right.\tag{A3}
\end{equation}
where $W_1$ and $W_2\in \mathbb{R}^{d\times 2d}$, $W_{3}$ and $W_{3}^{'} \in \mathbb{R}^{d\times d_w}$ are learnable parameters.

We implement the ablation study. As shown in Table~\ref{table4}, \textbf{CAG-DualQ} performs worse than \textbf{CAG}. It is interpretable. As illustrated in Fig.~\ref{figure8}, the Y-axis of the matrix $A^{(t)}$ marks the receiving nodes, and the X-axis denotes the distributing nodes. To infer an exact answer, for a node, we use the question command $q_w^{(t)}$ to activate its neighbors. In other words, the $i$-th row of the matrix $A_i^{(t)}$ calculates the correlation weights of node ${\cal N}_i^{(t)}$ and its neighbors $\{{\cal N}_j^{(t)} \}$ under the only once guidance of $q_w^{(t)}$. It is reasonable to introduce the question cue on one node side of $A^{(t)}$. 

\begin{table}
\centering
\setcounter{table}{3}
\fontsize{9.5}{10}\selectfont
\resizebox{0.48\textwidth}{!}{
\begin{tabular}{lccccc}
\hline
Model   & Mean$\downarrow$ & MRR$\uparrow$ & R@1$\uparrow$ & R@5$\uparrow$ & R@10$\uparrow$ \\
\hline
CAG-DualQ    & 3.79 & 67.19  & 54.16 & 83.44   & 91.32 \\
CAG        &\textbf{3.75} &\textbf{67.56} &\textbf{54.64} &\textbf{83.72} &\textbf{91.48}\\
\hline
\end{tabular}}
\caption{Ablation studies of different adjacent correlation matrix learning strategies on VisDial val v0.9.}
\label{table4}
\end{table}

\begin{figure}
\centering
\setcounter{figure}{7}
  \includegraphics[width=6.2cm]{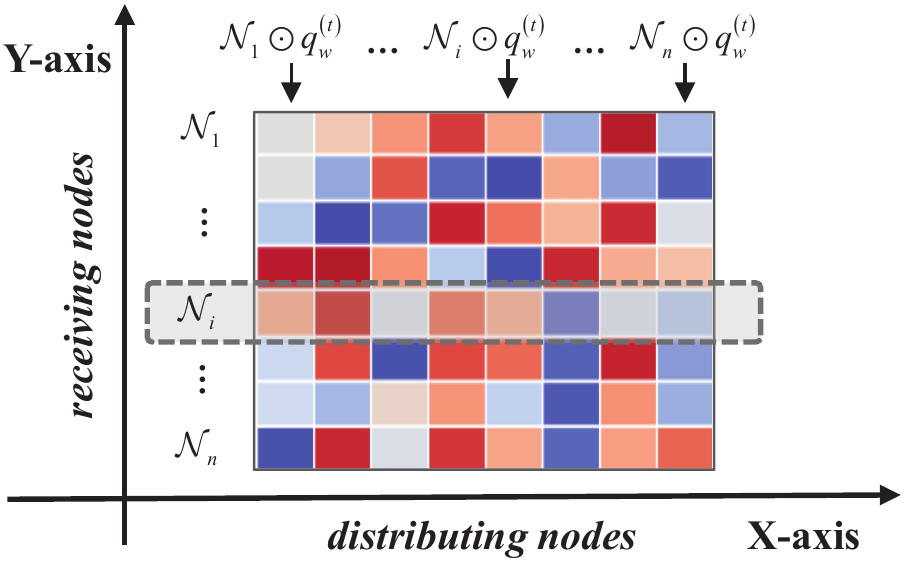}\\
  \caption{A schematic diagram of adjacent correlation matrix learning.}
  \label{figure8}
\end{figure}

\begin{figure*}
\centering
\setcounter{figure}{8}
  \includegraphics[width=17cm]{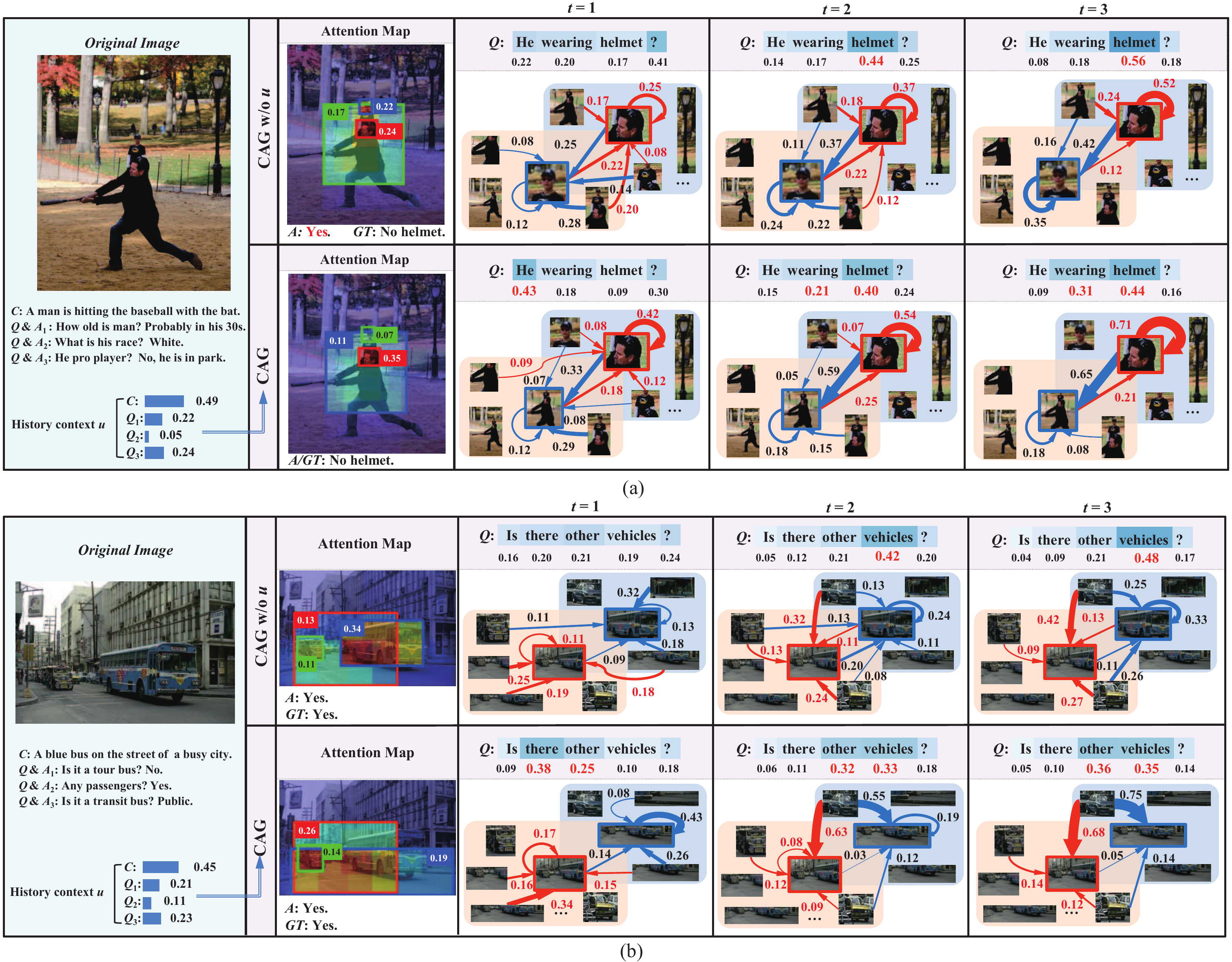}\\
  \caption{Qualitative results of \textbf{CAG} and \textbf{CAG w/o \emph{u}}. Observing the graph attention map overlaying each image, the bounding boxes with the attention scores correspond to the top-3 relevant object nodes in the final graph. We pick out the top-2 objects to display the dynamic graph inference. \textbf{CAG} and \textbf{CAG w/o \emph{u}} can refer to different top-2 objects. Without history context $u$, the agent could misunderstand the dialogue topic, and the visual reference cannot be solved well.}
  \label{figure9}
\end{figure*}
\begin{figure*}
\centering
\setcounter{figure}{9}
  \includegraphics[width=17cm]{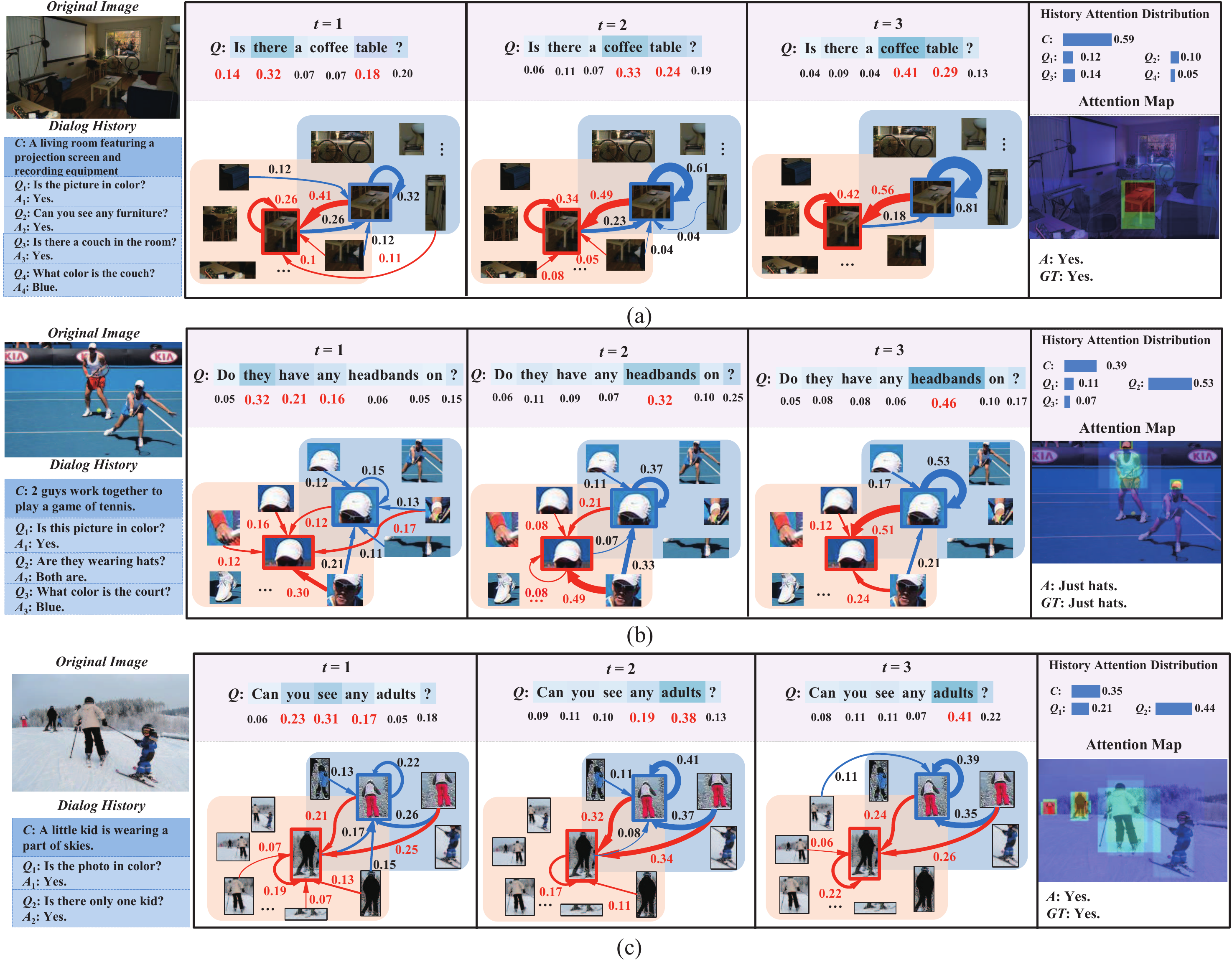}\\
  \caption{ Additional visualization examples on VisDial 0.9. In these attention maps, {\color{red}red} and {\color{blue}blue} bounding boxes correspond to the top-2 attended object nodes in the graph attention learning, respectively. }
  \label{figure10}
\end{figure*}
\subsection{Additional Qualitative Results}
\label{sec:s3}
\subsubsection{Qualitative results of CAG vs. CAG w/o \emph{u}}
As the ablation study shown in Table~\ref{table1}, the performance of \textbf{CAG w/o \emph{u}} drops a lot compared to \textbf{CAG}. Here, we provide explainable qualitative results in Fig.~\ref{figure9} to further validate the effectiveness of history-related context $u$ in \textbf{CAG}. There are two different examples. As shown in Fig.~\ref{figure9} (a), for question $Q$: ``He wearing helmet?'', \textbf{CAG w/o \emph{u}} directly locates words ``he'' (two people) and ``helmet'' in the image, and then the object ``helmet'' infers to a wrong ``he" (the man who wears the helmet), while \textbf{CAG} consistently attends on the correct ``he" (the subject ``man'' who is hitting the baseball with the bat in the previous dialogue). Besides, as shown in Fig.~\ref{figure9} (b), although \textbf{CAG w/o \emph{u}} infers the correct answer, but we observe that there is much more reasonable inference using \textbf{CAG} than \textbf{CAG w/o \emph{u}}. For the question $Q$: "Is there other vehicles?", \textbf{CAG} does not attend the bus in the center of picture and devote to searching other vehicles, while \textbf{CAG w/o \emph{u}} focuses on all the vehicles.

In a nutshell, \textbf{CAG w/o \emph{u}} is accustomed to attend all the objects appeared in the question, while \textbf{CAG} tries to ground the relevant objects discussed in the entire dialogue. If without the history reference, the dialogue agent can not perform the pronoun explanation ($e.g.,$ the visual grounding of ``he", ``it" and ``there", \emph{ect.}) well, and then the subsequent iterative inferences are affected. Therefore, the history-related context $u$ is necessary for the visual-textual co-reference reasoning in our solution.
\subsubsection{Additional qualitative results of visual-reference}
We provide additional four visualization results in Fig.~\ref{figure10}. These qualitative results also demonstrate that CAG has interpretable textual and visual attention distribution, reliable context-aware graph learning, and reasonable inference process.

\end{document}